%% file: main.tex
\documentclass[10pt,twocolumn,letterpaper]{article}

 \usepackage{cvpr}              
\usepackage{algorithm}
\usepackage{algorithmic}
\usepackage{placeins}                              
\input{preamble}

\definecolor{cvprblue}{rgb}{0.21,0.49,0.74}
\usepackage[pagebackref,breaklinks,colorlinks,allcolors=cvprblue]{hyperref}
\usepackage{float}

\def\paperID{7420} 
\def\confName{CVPR}
\def\confYear{2025}

\title{LLaVA-MR: Large Language-and-Vision Assistant for Video Moment Retrieval}

\author{Weiheng Lu\\
Peking University\\
\and
Jian Li${^{*}}$\\
Tencent Youtu\\
\and
An Yu\\
University at Albany\\
\and
Ming-Ching Chang\\
University at Albany\\
\and
Shengpeng Ji\\
Zhejiang University\\
\and
Min Xia\\
Peking University\\
}

\begin{document}
 \maketitle

 {
\renewcommand{\thefootnote}{\fnsymbol{footnote}}
\footnotetext[1]{ corresponding authors. swordli@tencent.com}
}
 
 \input{sec/0_abstract}    
 \input{sec/1_intro}
\input{sec/2_relatedwork}\input{sec/3_methods}
\input{sec/4_experiment}
 \input{sec/5_con}

 {
     \small
     \bibliographystyle{ieeenat_fullname}
     \bibliography{main}
 }
\input{sec/X_suppl}
\end{document}

%% file: preamble.tex
\usepackage{xcolor}

\usepackage{enumitem}

%% file: sec/0_abstract.tex
\begin{abstract}


Multimodal Large Language Models (MLLMs) are widely used for visual perception, understanding, and reasoning. However, long video processing and precise moment retrieval remain challenging due to LLMs’ limited context size and coarse frame extraction. We propose the Large Language-and-Vision Assistant for Moment Retrieval (LLaVA-MR), which enables accurate moment retrieval and contextual grounding in videos using MLLMs. LLaVA-MR combines Dense Frame and Time Encoding (DFTE) for spatial-temporal feature extraction, Informative Frame Selection (IFS) for capturing brief visual and motion patterns, and Dynamic Token Compression (DTC) to manage LLM context limitations. Evaluations on benchmarks like Charades-STA and QVHighlights demonstrate that LLaVA-MR outperforms 11 state-of-the-art methods, achieving an improvement of 1.82\% in R1@0.5 and 1.29\% in mAP@0.5 on the QVHighlights dataset. Our implementation will be open-sourced upon acceptance.

\end{abstract}

%% file: sec/1_intro.tex
\section{Introduction}
\label{sec:intro}
\begin{figure}[t] 
\centerline{
    \includegraphics[width=\linewidth]{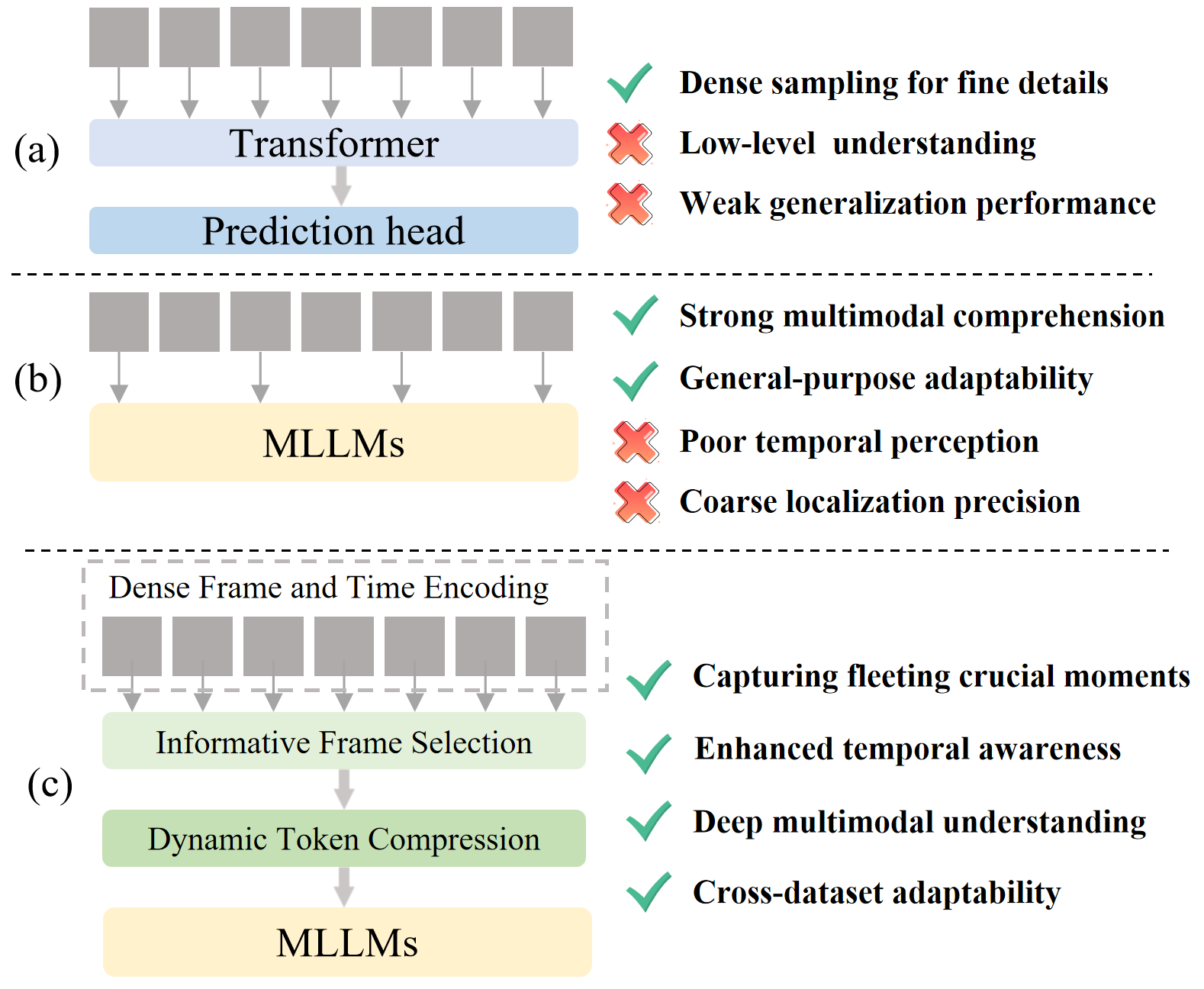} 
\vspace{-2mm}
}    
\caption{Advantages and Comparison of LLaVA-MR over prior work. (a) Traditional transformer-based methods. (b) Previous MLLMs-based methods. (c) Our LLaVA-MR.} 
    \label{fig:tease_image} 
\vspace{-2mm}    
\end{figure}
Large language models (LLMs) are widely recognized for their strong semantic understanding capabilities. Multimodal large language models (MLLMs)~\cite{jin2024efficientmllm, li2024surveybenchmarksmultimodallarge}, which extend LLMs to process multimodal data like video, have shown superior performance in tasks such as video description and visual question answering. They are also effective for specific visual tasks like object detection and visual grounding. However, using MLLMs for {\em moment retrieval} in long-duration videos is still a major challenge, as it requires a deep understanding of visual changes and precise integration of temporal information.

Video Moment Retrieval (MR) seeks to locate specific time segments within an untrimmed video based on a natural language query describing an event, requiring the model to effectively understand the video content and precisely capture the transient details relevant to the query. As illustrated in Fig.~\ref{fig:tease_image}(a), previous approaches, such as Moment-DTER~\cite{lei2021Moment-DETR}, CGDETR~\cite{moon2024CGDETR}, UniLoc~\cite{yan2023UniLoc}, and UniMD~\cite{zeng2024UniMD}, have primarily tackled this task by utilizing transformer structures to fuse image-text features, with task-specific prediction heads producing a fixed set of moment candidates.
However, these methods are limited by their reliance on frame-level feature extraction, which limits understanding, and the design of prediction heads reduces robustness when applied to new data.
Mr. BLIP~\cite{meinardus2024Mr.Blip} represents a breakthrough by framing moment retrieval as an open-ended sequence-to-sequence problem with a generative MLLM, offering strong multimodal understanding and versatile adaptability, as shown in Fig.~\ref{fig:tease_image}(b). However, this method is constrained by the limited context length of LLMs to process long sequences, which forces it to use sparse frame sampling, leading to poor temporal perception and coarse localization precision.

We identify two key challenges limiting the effectiveness of MLLMs in moment retrieval: (1) improving their temporal awareness and perception to enable more precise alignment of video frames with time encoding, and (2) ensuring more accurate localization of brief moments within long videos. Due to limited context size, MLLMs often rely on sparse sampling,  which reduces their ability to capture transient visual details in longer videos.

To address these challenges, we propose the {\bf Large Language-and-Vision Assistant for Moment Retrieval (LLaVA-MR)}, which optimizes MLLMs for moment retrieval tasks by enhancing their ability to capture critical, brief moments within long video frame sequences, as illustrated in Fig.~\ref{fig:tease_image}(c). 
To enhance temporal awareness and achieve higher-precision temporal perception, we apply Dense Frame and Time Encoding (DFTE) to extract fine-grained spatial and temporal features. To precisely locate brief moments in long videos, we developed an Informative Frame Selection (IFS) strategy to capture short-lived visual and motion patterns, and applied Dynamic Token Compression (DTC) to reduce sequence length while preserving essential information.
This enables MLLMs to retrieve moments over extended video frame sequences effectively. Our approach achieves state-of-the-art results on widely used benchmarks, including Charades-STA~\cite{gao2017tall} and QVHighlights~\cite{lei2021detecting}. Notably, it achieves an improvement of 1.82\% in R1@0.5 and 1.29\% in mAP@0.5 on the QVHighlights dataset.

In summary, our main contributions are the following:
\begin{itemize}[leftmargin=10pt] \itemsep -.1em
\item We propose LLaVA-MR, which applies MLLMs to the moment retrieval task, enhancing temporal awareness and effectively capturing critical, transient information in long videos.
\item We introduce Dense Frame and Time Encoding (DFTE) to extract fine-grained spatial and temporal features, Informative Frame Selection (IFS) to capture critical dynamic moments, and Dynamic Token Compression (DTC) to compress redundant information.
\item Our method achieves state-of-the-art performance on the moment retrieval task across both the Charades-STA and QVHighlights datasets, demonstrating a greater advantage on more complex, longer videos.
\end{itemize}

%% file: sec/2_relatedwork.tex
\section{Related Work}
\label{sec: related work}

\subsection{Moment Retrival}


Moment Retrieval (MR) aims to identify video segments that match natural language queries, with Temporal Action Detection (TAD)~\cite{lin2020dbg} serving as a sub-task to localize and classify actions within untrimmed videos. Early MR methods, such as TALL~\cite{gao2017tall}, relied on sliding window techniques for temporal localization, though these were computationally expensive. Hendricks~\cite{anne2017localizing} improved efficiency by introducing attention-based boundary prediction. Recent architectures, such as Semantic Conditioned Dynamic Modulation by Yuan~\cite{yuan2019semantic} and the Moment Alignment Network (MAN) by Zhang~\cite{zhang2019man}, dynamically adjust video features based on query semantics and iteratively align video and text using graph adjustments. To capture temporal dependencies, Zhang~\cite{zhang2020learning} introduced a 2D Temporal Adjacent Network employing 2D convolutions. In TAD, multi-stage CNNs~\cite{shou2016temporal}, TCNs~\cite{lea2017temporal}, and BSN~\cite{lin2018bsn} improved boundary proposals, evolving into structured models like SSN~\cite{zhao2017temporal} and GCNs~\cite{zeng2019graph}. Transformer-based models, such as Girdhar’s attention-based model~\cite{girdhar2019video} and ActionFormer by Zhang~\cite{zhang2022actionformer}, along with multi-scale techniques by Dai~\cite{dai2021ctrn,dai2022ms} and enriched context modeling by Zhu~\cite{zhu2021enriching} and Sardari~\cite{sardari2023pat}, have further advanced temporal modeling within MR.
Additionally, transformer-based models have gained prominence, with Liu~\cite{liu2022umt} introducing Unified Multi-Modal Transformers (UMT) for joint moment retrieval and highlight detection. Mun~\cite{mun2020local} improved temporal grounding by integrating local-global video-text interactions through multi-level architecture. These methods, while promising, often suffer from annotation bias due to the structure of the prediction head, which impacts their generalization. LLaVA-MR, built on MLLMs, not only offers greater general-purpose versatility but also demonstrates improved performance due to its deeper multimodal understanding.
\begin{figure*}[t]
\centerline{
    \includegraphics[width=1\linewidth]{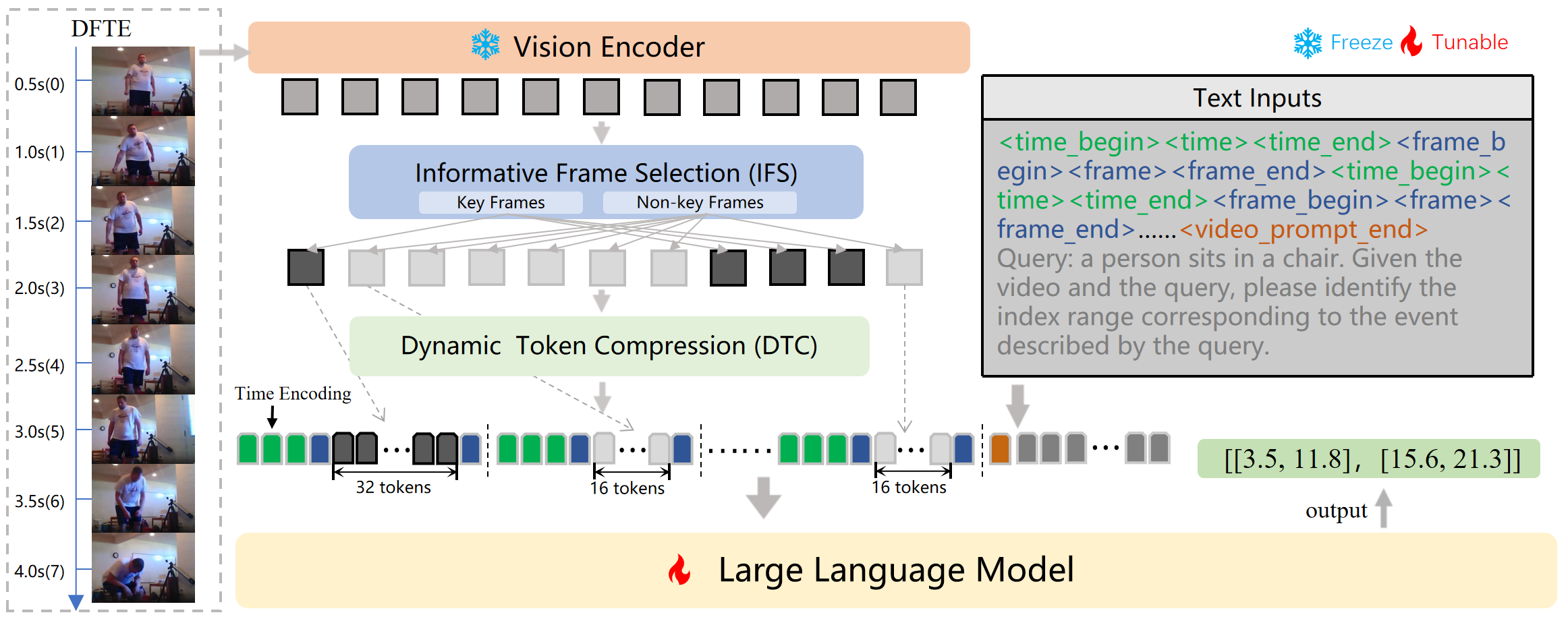} 
\vspace{-2mm}    
}
\caption{LLaVA-MR model overview. We leverage a pretrained MLLM such as BLIP-2\cite{li2023Blip2}. Our model primarily consists of Dense Frame and Time Encoding (DFTE), Informative Frame Selection (IFS), and Dynamic Token Compression (DTC).}
\label{fig:architecture} 
\vspace{-2mm}
\end{figure*}

\subsection{Multimodal Large Language Models}
Building on the success of Large Language Models (LLMs), researchers are advancing Multimodal Large Language Models (MLLMs) to enhance cross-modal understanding by integrating visual, auditory, and textual data. Some methods~\cite{alayrac2022flamingo, awadalla2023OpenFlamingo, li2023MIMIC-IT, zhang2024LLaMA-Adapter} enhance LLMs by incorporating additional parameters, such as gated cross-attention layers or adapter layers, to enable them to process multimodal inputs.
Other approaches~\cite{dai2023InstructBLIP, liu2023LLaVA, zhu2023MiniGPT-4} utilize modules like projection layers or Q-Formers to map the outputs from visual encoders into the input space of LLMs. Video LLMs~\cite{li2024VideoChat, zhang2023Video-LLaMA, maaz2024Video-ChatGPT, luo2023Valley} extend MLLMs for video tasks, primarily employing projection layers or Q-Formers for mapping visual tokens.  Mr. BLIP~\cite{meinardus2024Mr.Blip}, leveraging MLLMs, has achieved state-of-the-art results in moment retrieval with better semantic understanding and generalization. However, Mr. BLIP’s sparse sampling, constrained by the LLM’s window size, tends to miss fleeting visual details. Building upon Mr. BLIP, we capture more granular visual and temporal information with Dense Frame and Time Encoding (DFTE), while using Informative Frame Selection (IFS) and Dynamic Token Compression (DTC) to preserve critical dynamic moments and compress redundant data.

%% file: sec/3_methods.tex
\section{Method}


Our work enhances MLLMs for moment retrieval by improving temporal perception and capturing critical moments in long video sequences. 
$\S$~\ref{architecture} provides an overview of the LLaVA-MR architecture.
$\S$~\ref{sec:DFTE} details  the \textbf{Dense Frame and Time Encoding (DFTE)}, which extracts fine-grained visual features with time encoding for improved temporal awareness $\S$~\ref{sec:IFS} covers \textbf{Informative Frame Selection (IFS)} for capturing critical, brief moments within long video frame sequences. Finally, $\S$~\ref{sec:DTC} describes how the \textbf{Dynamic Token Compression (DTC)} reduces sequence length while preserving essential information. $\S$~\ref{sec: train and inference} outlines the model training and inference process.

\subsection{LLaVA-MR Architecture}
\label{architecture}
Our model architecture is illustrated in Fig.~\ref{fig:architecture}. LLaVA-MR starts with a frozen image encoder that extracts visual features from all sampled frames, generating a sequence of image embeddings defined as \( \mathbf{F}^v = [\mathbf{f}_1^v, \dots, \mathbf{f}_N^v] \in \mathbb{R}^{N \times P \times D_0} \), where \( N \) represents the number of video frames, \( P \) is the number of patches, and \( D_0 \) is the visual feature dimension. We then apply Informative Frame Selection (IFS) to evaluate visual changes between frames, categorizing them as {\em key} or {\em non-key}. {\em Key} frames capture dynamic variations, often at event boundaries, while {\em non-key} frames show minimal change and carry more redundant information. Next, Q-Former, a query-based transformer, refines each frame embedding by extracting relevant information. {\em Key} frame embeddings are denoted as \( \mathbf{K}^q = [\mathbf{f}_{i_1}^{q}, \mathbf{f}_{i_2}^{q}, \dots, \mathbf{f}_{i_k}^{q}] \in \mathbb{R}^{k \times Q \times D_1} \), and {\em non-key} frame embeddings are denoted as \( \mathbf{N}^q = [\mathbf{f}_{j_1}^{q}, \mathbf{f}_{j_2}^{q}, \dots, \mathbf{f}_{j_{N-k}}^{q}] \in \mathbb{R}^{(N-k) \times Q \times D_1} \), where \( k \) is the number of {\em key} frames, \( Q \) represents the number of queries, and \( D_1 \) is the feature dimension after Q-Former. For {\em non-key} frames, we apply Dynamic Token Compression (DTC) within \( \mathbf{N}^q \) to reduce sequence length while preserving essential information. Both \(\mathbf{K}^q \) and the compressed \( \mathbf{N}^q \) are projected into the language space, forming a unified embedding space,  which is defined as \( \mathbf{F}^L = [\mathbf{f}_1^L, \mathbf{f}_2^L, \dots, \mathbf{f}_N^L] \). 

We then construct an interleaved multimodal input sequence consisting of time tokens \( t_i \), frame embeddings \( \mathbf{f}_i^L \in \mathbf{F}^L \), a query \( q \), and a task prompt \( p \). We position each time token directly before its corresponding frame embeddings, using special tokens \( T_B \), \( T_E \), \( F_B \), and \( F_E \) to mark the start and end of the time and frame segments. Our interleaved multimodal sequence is structured as:
\begin{equation}
x = [T_B, t_1, T_E, F_B, \mathbf{f}_1^L, F_E, \dots, q, p]
\end{equation}
where \( t_1 \) represents the first time token, and \( \mathbf{f}_1^L \) represents the first frame embedding.

Finally, this sequence is fed into the LLM to predict the time segments relevant to query $q$. Following the approach of Mr. BLIP~\cite{meinardus2024Mr.Blip}, the LLM is tasked with predicting the sequence of potentially multiple relevant moments, each defined by a start and end time in seconds. The prediction follows the format of a nested list of moments: 
\(
y = [[t_1^{\text{start}}, t_1^{\text{end}}], [t_2^{\text{start}}, t_2^{\text{end}}], \dots]
\).

\subsection{Dense Frame and Time Encoding}
\label{sec:DFTE}
We use Dense Frame and Time Encoding (DFTE) to capture detailed visual information and associate each frame with its corresponding time, enhancing the temporal awareness of MLLMs.
Previous MLLM-based methods primarily rely on sparse sampling, which risks losing brief yet essential details. To address this, we increase the number of sampled frames, effectively capturing dense visual information and ensuring that critical transitions are not missed.

To improve MLLMs' temporal awareness, we propose choosing time encoding based on the frame sampling rate, using either {\em relative frame indices} or {\em timestamps} in seconds to avoid confusion from split tokens in large or decimal values.
We define the {\em sampling rate} $R_{frames}$ as the ratio of the total sampled frames to the video duration ($R_{frames} = \frac{N}{T}$), where $N$ is the number of sampled frames, and $T$ is the video length. Two design scenarios are considered: 
(1) When $R_{frames} \geq 1$, using relative frame indices as time tokens works best. Rounding timestamps in seconds to the nearest integer can create overlaps, which may confuse the LLM. For example, timestamps like 2.8 and 3.3 seconds would both round to 3, causing confusion for the LLM. In contrast, frame indices provide a clearer, non-overlapping representation.
(2) When $R_{frames} < 1$, where frames are sampled more sparsely, and the time interval between adjacent indices is relatively large. Here, using timestamps as time tokens provides a clearer temporal reference. $\S$~\ref{sec:ablation} will discuss additional temporal representations and explore the effect of different time representations as time tokens regarding model performance.
\subsection{Informative Frame Selection}
\label{sec:IFS}
Dense sampling captures detailed visual information but adds redundancy, making it crucial to isolate key visual and motion patterns. Brief transitions, like setting down a cup before picking up a phone, often hold more useful information, which is unevenly distributed across frames. To address this, we introduce the Informative Frame Selection (IFS) module, which captures brief yet crucial visual and motion variations.

In the IFS module, We categorize frames into two types: \emph{key} frames and \emph{non-key} frames. \emph{Key} frames capture notable changes, often marking important event transitions and carrying valuable information. In contrast, \emph{non-key} frames show minimal change, typically representing ongoing activities already covered by previous frames. 
So how do we distinguish between {\em key} frames and {\em non-key} frames? For \( \mathbf{F}^v \) obtained from the image encoder, we first compute the change between adjacent frames by subtracting each frame from the preceding one, denoted as \( \Delta \mathbf{F}^v = [\Delta \mathbf{f}_1^v, \dots, \Delta \mathbf{f}_N^v] \in \mathbb{R}^{(N \times P \times D_0)} \), where each component represents the frame-to-frame variation across patches.
For each \( \Delta \mathbf{f}_i^v \) in \(\Delta \mathbf{F}^v\), we then compute its L2 norm to obtain \( \mathbf{d} = [d_1, d_2, \dots, d_N] \). To further reduce the impact of video jitter, we apply a Gaussian filter to \( \mathbf{d} \), resulting in the adjacent feature distance \( \hat{\mathbf{d}} \), where each element \( \hat{d}_i \) represents the change of the \( i \)-th frame relative to the previous one: 
\vspace{-2mm}
\begin{equation}
\hat{d}_i = 
\begin{cases}
\max(\mathbf{d}), & \text{if } i = 0 \\
\mathcal{G}(\|\Delta \mathbf{f}_i^v\|_2), & \text{if } i > 0
\end{cases}
\vspace{-2mm}
\end{equation}
where, \( \|\Delta \mathbf{f}_i^v\|_2 \) denotes the L2 norm of the frame difference matrix \( \Delta \mathbf{f}_i^v \), and \( \mathcal{G} \) represents the Gaussian filter applied to smooth \( \mathbf{d} \). Then, we select the top \( k \) frames with the highest \( \hat{\textbf{d}} \) values as {\em key} frames, while the remaining frames are classified as {\em non-key} frames. We can define this formally as follows:
\begin{equation}
\left\{
\begin{aligned}
    \mathbf{K}^v &= [\mathbf{f}_{i_1}^v, \dots, \mathbf{f}_{i_k}^v], \quad i_j \in \text{top}_k(\hat{\mathbf{d}}) \\
    \mathbf{N}^v &= [\mathbf{f}_{j_1}^v, \dots, \mathbf{f}_{j_{N-k}}^v], \quad j_l \notin \text{top}_k(\hat{\mathbf{d}})
\end{aligned}
\right.
\end{equation}
where \( \mathbf{K}^v \) represents the image embeddings of {\em key} frames, and \( \mathbf{N}^v \) represents image embeddings of {\em non-key} frames.
We will discuss the selection of the total frame number \(N\) and the setting of $k$ in $\S$~\ref{sec:ablation}.

\begin{figure}[t]
\centerline{
    \includegraphics[width=1\linewidth]{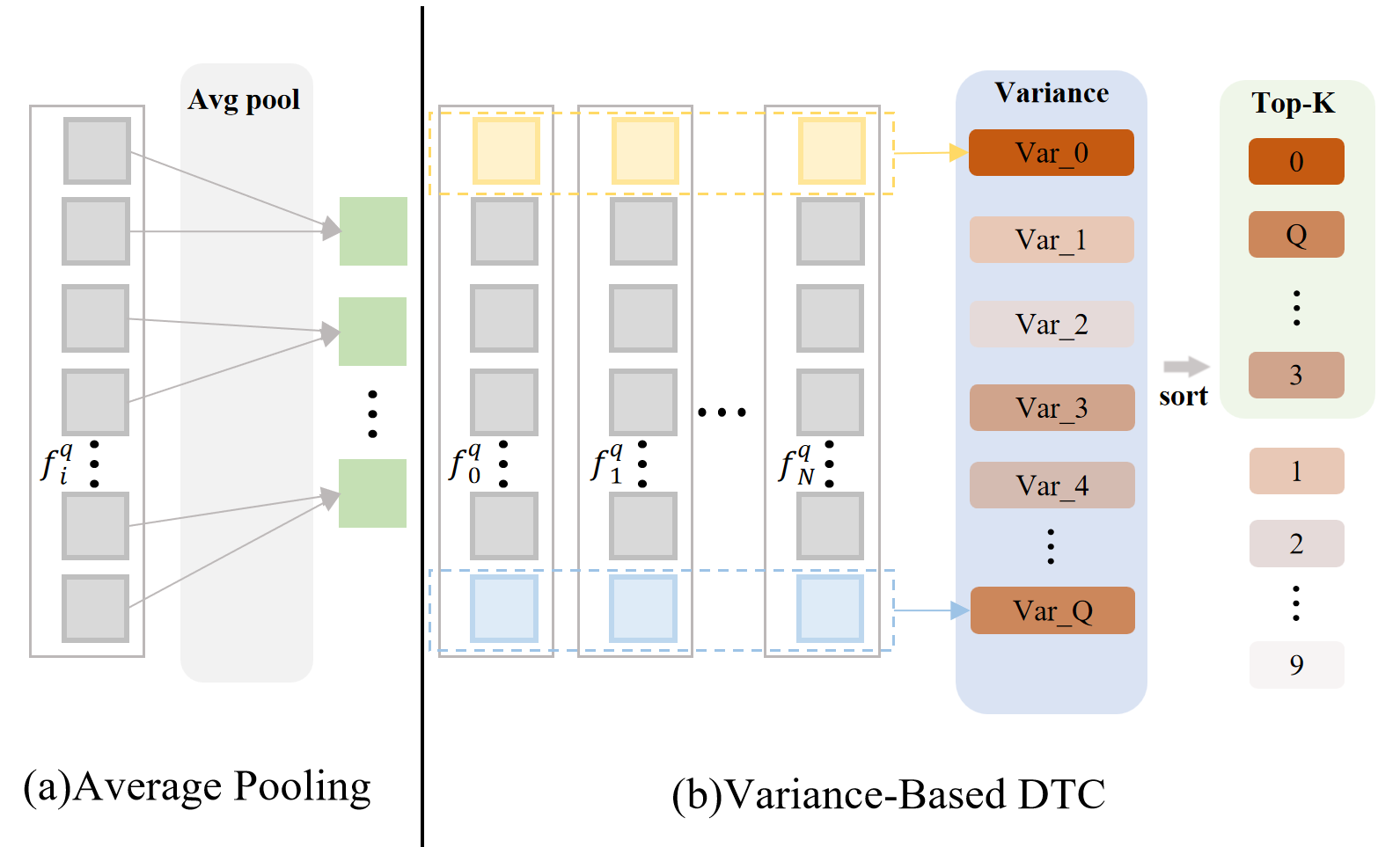} 
\vspace{-2mm}    
}
\caption{Two compression methods in \textbf{DTC}. (a) Average Pooling reduces tokens by averaging feature values within a window of 2, where $f_i^q \in N^q$. (b) Variance-Based DTC calculates the variance of each query in the Q-Former across all frames (where Q is the number of queries), sorts queries by variance, and selects the top half of queries to retain tokens focusing on dynamic content.}
\label{fig:DTC} 
\vspace{-2mm}
\end{figure}
\subsection{Dynamic Token Compression}
\label{sec:DTC}
As highlighted in $\S$~\ref{sec:IFS}, {\em non-key} frames typically exhibit fewer dynamic changes, often representing the repetitive or static continuation of an event. On the other hand, when frame sequences are too long, this not only reduces the performance of the LLM but also leads to significant resource consumption. Therefore, we employ the Dynamic Token Compression (DTC) module to compress {\em non-key} frame embeddings, thereby reducing the sequence length while retaining valuable information.

After dividing the frames into {\em key} and {\em non-key} frames, we input \(\mathbf{K}^v \) and \( \mathbf{N}^v \) into the Q-Former to refine the embeddings. The refined embeddings for {\em key} and {\em non-key} frames are \( \mathbf{K}^q \in \mathbb{R}^{k \times Q \times D_1} \) and \( \mathbf{N}^q \in \mathbb{R}^{(N-k) \times Q \times D_1} \) ($\S$~\ref{architecture}). 
We propose two compression methods: average pooling, and Variance-Based Dynamic Token Compression (Variance-Based DTC). As shown in Fig.~\ref{fig:DTC}, average pooling reduces tokens by averaging feature values within a given window. In contrast, the variance-based DTC method aims to retain Q-Former queries that capture dynamic content in the video frames. Considering the idea that each query in the Q-Former focuses on different regions or aspects of the visual content, some queries, such as those tracking static elements like backgrounds, may offer little to no guidance in predicting relevant results. 
Based on this perspective, we calculate the variance of each query across frames, retaining only those queries with the highest variance—queries more likely to capture moving or dynamic aspects of the scene.
$\S$~\ref{sec:ablation} will discuss the impact of different token compression methods on our model performance.

\subsection{Model Training and Inference}
\label{sec: train and inference}
During training, we fine-tune our generative MLLM to tackle moment retrieval by reframing it as an open-ended, sequence-to-sequence task in natural language. The model is trained using a maximum likelihood objective. For each input \( x \) and target \( y \), the loss is minimized as:
\vspace{-2mm}
\begin{equation}
\mathcal{L}(x, y) = - \sum_{i=1}^{L-1} \log \, p_\theta(y_{i+1} \, | \, x, y_{1:i})
\vspace{-2mm}
\end{equation}
%
where \( L \) is the length of the target sequence. We employ the LoRA~\cite{hu2021lora} parameter-efficient fine-tuning approach to effectively adapt with minimal additional parameters by injecting low-rank matrices into the model layers.

For inference, we sample frames and use the text decoder to generate the moment sequence, guided by the model’s likelihood with beam search. The final sequence is output as a Python-style list of predicted moments, with minor formatting adjustments made as needed.
Notably, when using relative frame indices as time encoding, we first record the mapping between these indices and actual timestamps, and then convert the output of the MLLM back to specific timestamps. This step is unnecessary when timestamps are used directly as the temporal representation.

%% file: sec/4_experiment.tex
\section{Experiment}
\label{sec: experiment}
This section outlines the experimental details and evaluates the effectiveness of our design choices. In $\S$~\ref{sec: Experiment Setup}, we first provide an in-depth description of our experimental setup and implementation details. In $\S$~\ref{sec: results}, We then present a comparison with the state-of-the-art methods and qualitative results from our approach. Finally, We explore the results of our ablation studies in $\S$~\ref{sec:ablation}. 

\subsection{Experiment Setup}
\label{sec: Experiment Setup}
\textbf{Benchmarks.} We validate our LLaVA-MR on the two most widely used video moment retrieval (MR) datasets:

Charades-STA~\cite{gao2017tall} comprises 9,848 videos with an average duration of 30.6 seconds. The dataset includes 16,128 annotations, with an average moment duration of 8.1 seconds and an average query length of 7.22 words. It is split into two subsets: training (12,408 annotations) and testing (3,720 annotations). 



QVHighlights~\cite{lei2021detecting} is a recent benchmark with 10,148 videos, each lasting 150 seconds. These videos are cropped from YouTube content and annotated with at least one query, averaging 11.3 words, to describe the relevant moments. The target moments have an average duration of 24.6 seconds. The dataset is divided into training, validation, and test sets, comprising 7,218, 1,150, and 1,542 queries, respectively. To ensure fair evaluation, the test set targets are kept hidden, and performance on the test split is only measurable through prediction submissions to the evaluation server.~\footnote{evaluation server: \url{https://codalab.lisn.upsaclay.fr/competitions/6937}}

\noindent\textbf{Metrics.}
Common metrics for Moment Retrieval (MR) include Recall@K, mean Average Precision (mAP), and mean Intersection over Union (mIoU), each evaluated at different IoU thresholds. Recall@K represents the percentage of the top-K predicted segments with an IoU above a given threshold (e.g., R1@0.5 is Recall@1 at 0.5 IoU). mIoU measures the average IoU across predictions, while mAP indicates mean precision at specific IoU thresholds, such as mAP@0.5 and mAP@0.75. In brief, R1@0.5 and R1@0.7 focus on the top-1 prediction’s recall at different IoU thresholds, mIoU measures average overlap, and mAP captures precision at varying IoUs.

\noindent\textbf{Implementation details.} 
Our experimental setup mainly follows the configurations of Mr. BLIP~\cite{meinardus2024Mr.Blip}. Considering the relatively small scale of the datasets, we employ LoRA-based parameter-efficient finetuning to train only 0.63\% of the total parameters in the model. Due to the output instability of LLMs, we mitigate minor formatting inconsistencies by applying post-processing heuristics to improve prediction accuracy.
Our optimizer is AdamW~\cite{loshchilov2019AdamW}, with an initial learning rate (LR) of 1e-8, linearly warmed up to 3e-4 over 10\% of iterations, followed by cosine LR decay. Frames are sampled randomly during training.

For Charades-STA, we sample 60 frames per video, training for up to 50 epochs with a batch size of 16 across 4 A100-80GB GPUs, totaling around 70 GPU hours. For QVHighlights Captions, we sample 80 frames per video, train up to 50 epochs with a batch size of 32 on 8 A100-80GB GPUs, and set gradient accumulation to 4, totaling about 33 GPU hours.

\subsection{Results}
\label{sec: results}

\noindent\textbf{Comparison to the State-of-the-Art.} To evaluate our approach for moment retrieval, we compare LLaVA-MR against SoTA methods on two widely used datasets: Charades-STA and QVHighlights. As summarized in Table \ref{tab:sota}, LLaVA-MR achieves the highest performance across all key metrics on both datasets. This not only demonstrates the superiority of our approach but also confirms its robustness across datasets with varying complexities and multiple evaluation metrics. 
Notably, on the QVHighlights dataset, which contains longer videos and requires a deeper understanding compared to the Charades-STA dataset, LLaVA-MR demonstrates a more substantial improvement. On the validation set, it achieves a \textbf{2\%} increase in R1@0.5, while on the test set, it outperforms other methods with gains of \textbf{1.82\%} in R1@0.5, \textbf{0.97\%} in R1@0.7, \textbf{1.29\%} in mAP@0.5, and \textbf{1.02\%} in mAP@0.75. These results underscore our model’s enhanced capability to handle moment retrieval tasks in longer, more complex videos.

\noindent\textbf{Inference Time}. We compared the inference times of LLaVA-MR and MR. BLIP \cite{meinardus2024Mr.Blip} with the same number of sampled frames. After averaging 10 tests on an A100-80G GPU, LLaVA-MR took 1.31s, while MR. BLIP took 1.57s. The addition of modules such as IFS was offset by the compression mechanism, resulting in faster inference.

\begin{figure*}[t] 
\centerline{
  \includegraphics[width=0.98\linewidth]{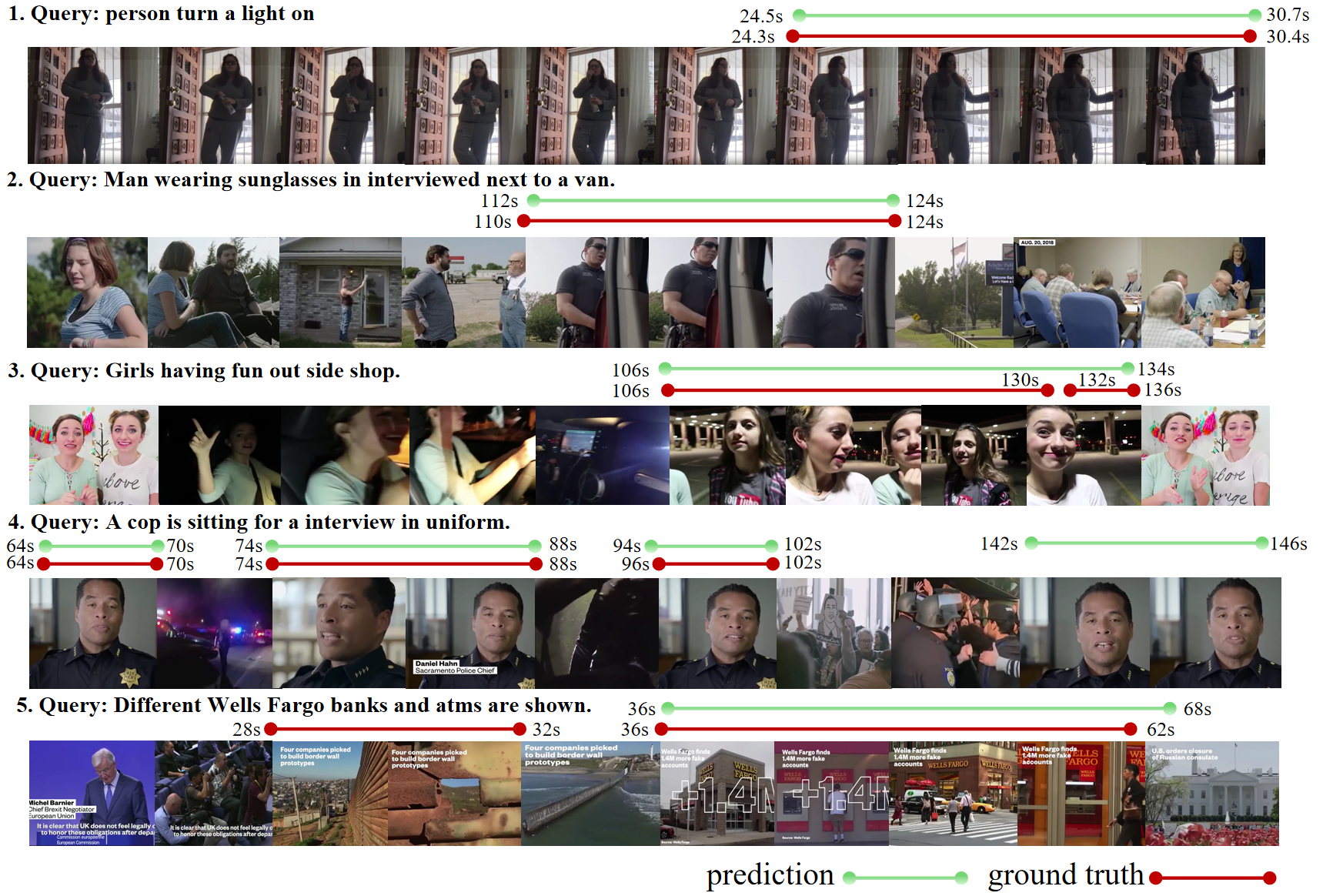} 
\vspace{-2mm}
}
\caption{Qualitative results on the Charades-STA and QVHighlights datasets, with ground truth segments for query events alongside highlighted predicted intervals.}
\label{fig:showcase} 
\vspace{-2mm}
\end{figure*}

\noindent\textbf{Qualitative Results}. Fig.~\ref{fig:showcase} visualizes qualitative results, with ground truth segments for query events alongside highlighted predicted intervals. In examples 1, 2, and 3, LLaVA-MR accurately identifies the segments matching the natural language queries, with predicted start and end times closely aligning with ground truth. This demonstrates the model’s strong ability to perceive temporal boundaries. In example 4, the model not only outputs the correct segment but also detects an additional relevant segment at the video’s end that was not annotated. Example 5 illustrates a failure case where the model predicts an obvious segment but misses another, less prominent one.


\begin{table}[t]
\caption{Comparison with State-of-the-Art on Charades-STA and QVHighlights Datasets. \textbf{CLIP*}: UnLoc~\cite{yan2023UnLoc-L} pretrains the backbone with action classification datasets Kinetics 400/700. \textbf{SF}: Slow-Fast backbone. 
\vspace{-2mm}
}
\label{tab:sota}
\centerline{
\setlength{\tabcolsep}{0.2mm}
\resizebox{1.05\linewidth}{!}{
\footnotesize
\begin{tabular}{lcccccc}
\toprule
Method & Backbone & mIoU & R1@0.5 & R1@0.7 & mAP@0.5 & mAP@0.75 \\
\midrule
\multicolumn{7}{c}{\textbf{QVHighlights Validation set}} \\
Moment-DETR\cite{lei2021Moment-DETR} & SF+CLIP & -- & 59.68 & 40.84 & -- & -- \\
EaTR~\cite{jang2023EaTR} & I3D & -- & 61.36 & 45.79 & 61.86 & 41.91 \\
QD-DETR~\cite{moon2023QD-DETR} & SF+CLIP & -- & 62.68 & 46.66 & 62.23 & 41.82 \\
UnLoc-L~\cite{yan2023UnLoc-L} & CLIP* & -- & 66.10 & 46.70 & -- & -- \\
CG-DETR~\cite{moon2024CGDETR} & SF+CLIP & -- & 67.40 & 52.10 & 65.60 & 45.70 \\
Mr. BLIP~\cite{meinardus2024Mr.Blip} & BLIP-2 & -- & 76.13 & 63.35 & 69.39 & 55.78 \\
\textbf{LLaVA-MR} & BLIP-2 & \textbf{71.85} & \textbf{78.13} & \textbf{64.13} & \textbf{69.64} & \textbf{56.32} \\
\midrule
\multicolumn{7}{c}{\textbf{QVHighlights Test set}} \\
SeViLa~\cite{yu2023SeViLa} & BLIP-2 & -- & 54.50 & 36.50 & -- & -- \\
UniVTG~\cite{lin2023UniVTG} & SF+CLIP & -- & 58.86 & 40.86 & 57.60 & 35.59 \\
QD-DETR~\cite{moon2023QD-DETR} & SF+CLIP & -- & 62.40 & 44.98 & 62.52 & 39.88 \\
CG-DETR~\cite{moon2024CGDETR} & SF+CLIP & -- & 65.43 & 48.38 & 64.51 & 42.77 \\
Mr. BLIP~\cite{meinardus2024Mr.Blip} & BLIP-2 & -- & 74.77 & 60.51 & 68.12 & 53.38 \\
\textbf{LLaVA-MR} & BLIP-2 & -- & \textbf{76.59} & \textbf{61.48} & \textbf{69.41} & \textbf{54.40} \\
\midrule
\multicolumn{7}{c}{\textbf{Charades-STA Test set}} \\
Moment-DETR~\cite{lei2021Moment-DETR} & & -- & 53.63 & 31.37 & -- & -- \\
QD-DETR~\cite{moon2023QD-DETR} & SF+CLIP & -- & 57.31 & 32.55 & -- & -- \\
UniVTG~\cite{lin2023UniVTG} & SF+CLIP & 50.10 & 58.01 & 35.65 & -- & -- \\
CG-DETR~\cite{moon2024CGDETR} & SF+CLIP & 50.13 & 58.44 & 36.34 & -- & -- \\
UnLoc-L~\cite{yan2023UniLoc} & CLIP* & -- & 60.80 & 38.40 & -- & -- \\
UniMD+Sync~\cite{zeng2024UniMD} & -- & -- & 63.98 & 44.46 & -- & -- \\
InternVideo2-1B~\cite{wang2024InternVideo2} & -- & --& 68.36 & 45.03 & -- & -- \\
EaTR~\cite{jang2023EaTR} & I3D & -- & 68.47 & 44.92 & -- & -- \\
Mr. BLIP~\cite{meinardus2024Mr.Blip} & BLIP-2 & 58.63 & 69.31 & 49.29 & -- & -- \\
InternVideo2-6B~\cite{wang2024InternVideo2} & -- & -- & 70.03 & 48.95 & -- & -- \\
\textbf{LLaVA-MR} & BLIP-2 & \textbf{59.78} & \textbf{70.65} & \textbf{49.58} & \textbf{69.96} & \textbf{39.66} \\
\bottomrule
\end{tabular}
}}
\vspace{-2mm}
\end{table}

\subsection{Ablation Studies}
\label{sec:ablation}
To evaluate the contribution of each component in our approach, we conducted a comprehensive ablation study on the Charades-STA dataset. Table~\ref{tab:Ablation study} shows these results. The baseline (a) involves sparse sampling from the video and inputting it into the MLLM, achieving an R1@0.5 score of \textbf{65.48}, confirming the feasibility of using MLLM for the moment retrieval task. In experiment (b), we simply increased the number of sampled frames based on the baseline to capture more visual information, which led to an improvement of \textbf{+0.89} in R1@0.5. Experiment (c) builds upon (b) by optimizing temporal representation, enhancing the model's perception of temporal sequences, resulting in a further improvement of \textbf{+2.15} in R1@0.5. Finally, in experiment (d), we incorporated the IFS and DTC modules to capture crucial moment and compress redundant information, leading to a significant improvement of \textbf{+2.13} in R1@0.5.


\begin{table}[t]
\caption{Comprehensive ablation study on Charades-STA
\vspace{-2mm}
}
\label{tab:Ablation study}
\centerline{
\setlength{\tabcolsep}{0.3mm}
    \resizebox{\linewidth}{!}{%
    \begin{tabular}{cccc|ccc}
        \toprule
        &Dense Frame & Time Encoding & IFS\&DTC & R1@0.5 & R1@0.7 & mIoU \\
        \midrule
        (a)& &  &  & 65.48 & 42.71 & 55.17 \\
        (b)&\checkmark &  &  & 66.37 & 44.65  & 57.50 \\
        (c)&\checkmark & \checkmark & & 68.52 & 47.29 & 58.16 \\
        (d)&\checkmark & \checkmark & \checkmark & \textbf{70.65} & \textbf{49.58} & \textbf{59.78} \\
        \bottomrule
    \end{tabular}%
}}
\vspace{-2mm}
\end{table}
\begin{table}[t]
\caption{Ablation on Time Encoding. Evaluation on Charades-STA and QVHighlights.
\vspace{-2mm}
}
\label{tab:Time Representation}
\centerline{
\begin{tabular}{lccc}
\toprule
\textbf{Time Representation} & \textbf{R1@0.5} & \textbf{R1@0.7} & \textbf{mIoU} \\
\midrule
\multicolumn{4}{l}{\textbf{Charades-STA} Test Set} \\
Relative frame indices & \textbf{67.83} & \textbf{46.68} & \textbf{57.78}         \\
Timestamps & 66.37 & 44.65 & 57.50     \\
\midrule
\multicolumn{4}{l}{\textbf{QVHighlights} Validation Set} \\
Relative frame indices & 74.77 & 61.74 & 68.45         \\
Timestamps & \textbf{76.06} & \textbf{62.65} & \textbf{70.01}        \\
\bottomrule
\end{tabular}
}
\vspace{-2mm}
\end{table}

\noindent\textbf{Design of Time Encoding.}
Designing an effective time encoding method is essential for enabling MLLMs to associate each frame with its corresponding time. Common approaches include using frame indices or timestamps to indicate each frame’s position. Frame indices can be either relative, representing the order within the sampled sequence ({\em e.g.}, $1, 2, 3, \dots, 60$), or absolute, marking the frame’s position in the full video ({\em e.g.}, $1, 15, 30, \dots, 900$).
Since larger numbers are often split into multiple tokens during tokenization, leading to potential numerical confusion and lower performance for LLMs~\cite{meinardus2024Mr.Blip}, we focus on relative indices over absolute ones. Similarly, directly inputting decimal timestamps can cause tokenization issues; thus, rounding to the nearest integer is a common approach, which we apply as well when using timestamps.

We selected the Charades-STA and QVHighlights datasets to validate the hypothesis that different time representations should be chosen based on the sampling rate, \( R_{\text{frames}} \) (see $\S$~\ref{sec:DFTE}). In Table~\ref{tab:Time Representation}, we show the impact of using different time representations on the Charades-STA and QVHighlights datasets. For the Charades-STA dataset (with an average video length of 30 seconds, each video sampled at 60 frames, \( R_{\text{frames}} \approx 60/30 > 1 \)), we observed that using relative frame indices improved the R1@0.5 score by \textbf{+1.46} compared to using timestamps. We infer this occurs because when \( R_{\text{frames}} \geq 1 \), neighboring timestamps rounded to integers can become identical (e.g., 0.7 and 1.1 both round to 1), causing MLLMs to confuse the corresponding frames.
Hence, using incremental relative frame indices as the time encoding provides a clearer representation in this case.
Conversely, for the QVHighlights dataset (with an average video length of 150 seconds, each video sampled at 80 frames, \( R_{\text{frames}} \approx 80/150 < 1 \)), using frame indices as time tokens led to better performance, improving the R1@0.5 score by \textbf{+1.29}. We hypothesize that this happens because the time gap between sampled frames is relatively large. For example, a frame with a relative index of 1 could represent 0.2 or 1.4 seconds, leading to coarser time granularity and information loss. Timestamps, however, provide finer precision.

Additionally, to enhance the model's ability to distinguish between tokens representing temporal and frame-based information, we introduced special tokens: \texttt{<time\_begin>}, \texttt{<time\_end>}, \texttt{<frame\_begin>}, and \texttt{<frame\_end>}. We evaluated their impact on Charades-STA, with results presented in Table~\ref{tab:prompt design}, demonstrating their effectiveness in improving model performance.



\begin{table}[t]
\caption{Ablation evaluation of prompt design on Charades-STA. B/E are special tokens indicating the begin/end of time or frames.
\vspace{-2mm}
}
\label{tab:prompt design}
\centerline{
\begin{tabular}{lccc}
\toprule
\textbf{Input Format} & \textbf{R1@0.5} & \textbf{R1@0.7} & \textbf{mIoU} \\
\midrule
No special tokens & 67.83 & 46.68 & 57.78  \\
B/E for time  & 67.98  & 46.80 & 57.93 \\
B/E for frames  & 68.25 & 47.07 & 58.02  \\
B/E for both & \textbf{68.52} & \textbf{47.29} & \textbf{58.16} \\
\bottomrule
\end{tabular}
}
\vspace{-2mm}
\end{table}

\begin{table}[t]
\caption{Ablation Study on Different Token Compression Methods. Evaluation on Charades-STA.
\vspace{-2mm}
}
\label{tab:token_compression}
\centerline{
\begin{tabular}{lccc}
\toprule
\textbf{Compression Method} & \textbf{R1@0.5} & \textbf{R1@0.7} & \textbf{mIoU} \\
\midrule
No Compression & 68.52 & 47.29 & 58.16 \\
Average Pooling & 69.43 & 48.17 & 58.77  \\
Variance-Based DTC & \textbf{70.65} & \textbf{49.58} & \textbf{59.78} \\
\bottomrule
\end{tabular}
}
\vspace{-2mm}
\end{table}

\begin{figure}[t] 
\centerline{
    \includegraphics[width=\linewidth]{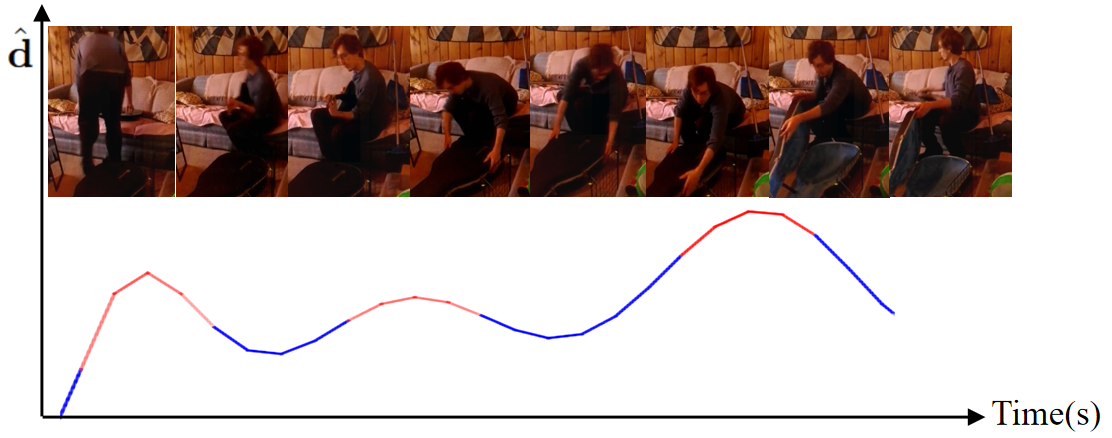} 
\vspace{-2mm}
}
\caption{Temporal correspondence between video frames and adjacent frame feature distance \( \hat{\mathbf{d}} \).}
    \label{fig:visualize} 
\vspace{-4mm}    
\end{figure}

\noindent\textbf{Design of Informative Frame Selection.} 
The Informative Frame Selection (IFS) module introduced in $\S$~\ref{sec:IFS} captures brief but critical visual and motion variations. This module classifies frames into {\em key} and {\em non-key} categories by calculating the adjacent feature distance \( \hat{\mathbf{d}} \). 
We selected a sample video and plotted \( \hat{\mathbf{d}} \), as shown in Fig.~\ref{fig:visualize}. It can be seen that \( \hat{\mathbf{d}} \) peaks when the person picks up or puts down the guitar, while remaining low during the guitar playing.
This supports the hypothesis that frames with more significant dynamic changes not only contain more valuable visual information but are also more likely to lie at event boundaries, while frames within event interiors remain relatively static.

We select the top-\( k \) frames with the highest change levels as {\em key} frames, with the remaining frames classified as {\em non-key}. The selection of the total number of frames \( N \) and top-\( k \) requires a balance between capturing sufficient visual information and managing the sequence length constraints of the LLM input.
As illustrated in Figure 4, we observe that when \( k \) is too small, excessive compression results in a significant loss of visual information, reducing model performance. Conversely, when \( k \) is too large, or even equals \( N \), redundant information increases the sequence length excessively, which also degrades model performance. Similarly, increasing \( N \) does not always improve performance; while \( N \) should ensure comprehensive sampling of visual variations, overly large values introduce longer sequence lengths and more redundant information, which can interfere with model performance. 
We found that \( N = 60 \) and \( k = 32 \) yield the best performance on Charades-STA. Following a similar approach, we set \( N = 80 \) and \( k = 32 \) for the QVHighlight dataset.

\begin{figure}[t] 
\centerline{
    \includegraphics[width=0.7\linewidth]{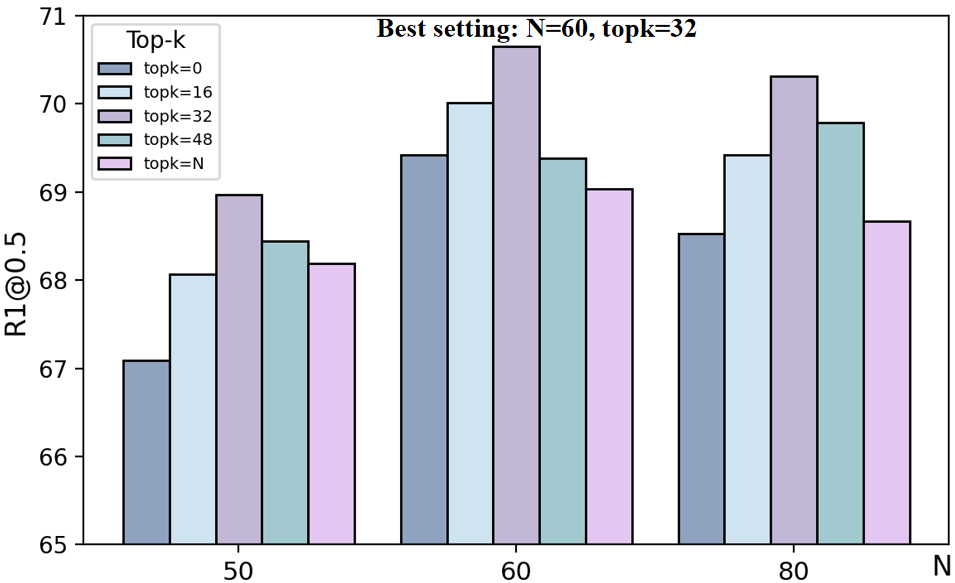} 
\vspace{-2mm}
}
\caption{Hyperparameter optimization results on the Charades-STA dataset to determine the optimal values for \( N \) and \( k \).} 
    \label{fig:NTopk} 
\vspace{-4mm}
\end{figure}


\noindent\textbf{Design of Dynamic Token Compression.} After dividing the frames into {\em key} and {\em non-key} frames, we employ the Dynamic Token Compression (DTC) module from $\S$~\ref{sec:DTC} to compress the features of non-key frames, thereby reducing the sequence length and eliminating redundancy. This compression strategy is designed to retain essential information while filtering out redundant data.


Table~\ref{tab:token_compression} shows the impact of different compression methods on the model's performance.
Without using any compression strategy, the model achieves a score of \textbf{68.52} on R1@0.5. 
We first applied average pooling to \( N^q \), reducing the token corresponding to each frame by half. This helped reduce redundancy and shorten the sequence length, achieving a score of \textbf{69.43} on R1@0.5. However, we infer that simply applying average pooling does not guarantee that more valuable information in \( N^q \) is preserved. This led us to the design of Variance-Based DTC, with an aim to retain Q-Former queries that capture dynamic content in the video frames. For example, in a video where a person is drinking water, a query focused on a static background element is unlikely to contribute meaningfully to understanding the action. Variance-Based DTC outperformed average pooling, improving R1@0.5 to \textbf{70.65}. We attribute this improvement to the fact that queries with higher variance are more likely to capture dynamic aspects of the scene, while lower-variance queries correspond to relative stable features across frames and can be more effectively complemented by features from other frames after compression.

%% file: sec/5_con.tex
\section{Conclusion}
We introduce LLaVA-MR, a novel approach that optimizes MLLMs for moment retrieval in long videos by enhancing their capability to capture brief but critical moments within extensive sequences. Our method combines Dense Frame and Time Encoding (DFTE), Informative Frame Selection (IFS), and Dynamic Token Compression (DTC), creating a robust framework for precise moment retrieval. Through comprehensive evaluations, LLaVA-MR demonstrates significant performance improvements, outperforming ten leading methods, especially on complex, long-video benchmarks like QVHighlights.



%% file: sec/X_suppl.tex
\clearpage
\setcounter{page}{1}
\maketitlesupplementary

\section{Additional Experiments}
\label{sec:Experiments}

We present additional experimental results, including a comparison with state-of-the-art methods on the ActivityNet Captions dataset and more findings from hyperparameter optimization.

\subsection{Comparison on ActivityNet Captions Dataset} 
ActivityNet Captions~\cite{krishna2017Anet} is a large-scale dataset consisting of 20,000 videos, each averaging 2 minutes in duration. It includes 72,000 segments, each paired with a human-written caption averaging 13.5 words. The dataset is organized into three subsets: training (37,421 segments), validation 1 (val\_1, 17,505 segments), and validation 2 (val\_2, 17,031 segments). In our experiments, the training set is used to train the model, while the val\_2 set is employed as the test set.

The results in Table~\ref{tab:sota_anet} show that LLaVA-MR achieves state-of-the-art performance on the ActivityNet Captions dataset, outperforming all baseline methods in both R1@0.5 and R1@0.7 metrics. Specifically, it attains an R1@0.5 score of \textbf{55.16}, surpassing Mr. BLIP\cite{meinardus2024Mr.Blip} by \textbf{1.24} points, highlighting the effectiveness of the LLaVA-MR architecture.

\subsection{Hyperparameter Optimization} 

{\bf Compression strategies and ratios:}
As discussed in the main paper, we investigated how different compression strategies impact model performance. Beyond the choice of strategy, the compression ratio is crucial. Each frame initially corresponds to 32 tokens, determined by the \textit{query\_number} of the Q-former, and the compressed token count is denoted as $T$. Low compression levels fail to reduce redundancy and alleviate sequence length constraints, while high compression levels lead to excessive information loss, degrading performance.

Table~\ref{tab:compression_ratios} shows that the model performs best with $T = 16$, achieving R1@0.5 of \textbf{70.65} and R1@0.7 of \textbf{49.58}. Performance drops significantly with both lower and higher compression levels, emphasizing the need for a balanced compression ratio.

\noindent{\bf Frame and top-k selection:}
We also studied the optimal number of frames ($N$) and top-$k$ values on the QVHighlights~\cite{lei2021Moment-DETR} dataset, aiming to balance sufficient visual information with the sequence length constraints of large language models (LLMs). Based on experiments on the Charades-STA~\cite{gao2017tall} dataset, where $N = 60$ and $k = 32$ yielded the best result, we adjusted these parameters for QVHighlights, which features significantly longer videos.
As shown in Table~\ref{tab:topk_n_settings_qvh}, the best performance was achieved with $N = 80$ and $k = 32$. Note, however, that these results were obtained without post-processing, so the identified configuration may not represent the best possible outcome overall.

\begin{table}[t]
\caption{Comparison with State-of-the-Art on ActivityNet Captions. \textbf{CLIP*}: UnLoc~\cite{yan2023UnLoc-L} pretrains the backbone with action classification datasets Kinetics 400/700. 
\vspace{-2mm}
}
\label{tab:sota_anet}
\centerline{
\footnotesize
\begin{tabular}{lcccccc}
\toprule
Method & Backbone & R1@0.5 & R1@0.7 \\
\midrule
DRN~\cite{zeng2020DRN} & -- & 45.45 & 24.36  \\
VLG-Net~\cite{soldan2021VLG-Net} & -- & 46.32 & 29.82  \\
UniLoc-B~\cite{yan2023UniLoc} & CLIP* & 48.00 & 29.70   \\
UnLoc-L~\cite{yan2023UniLoc} & CLIP* & 48.30 & 30.20  \\
GVL~\cite{wang2023GVL} & -- & 49.18 & 29.69 \\
Mr. BLIP~\cite{meinardus2024Mr.Blip} & BLIP-2 & 53.92 & 35.55 \\
\textbf{LLaVA-MR} & BLIP-2 & \textbf{55.16} & \textbf{35.68} \\
\bottomrule
\end{tabular}
}
\vspace{-2mm}
\end{table}

\begin{table}[t]
\caption{Evaluation of Different Compression Ratios on Charades-STA. $T$ is the number of tokens after compression, and the ratio is the number of compressed tokens divided by the original (32).}
\label{tab:compression_ratios}
\centering
\begin{tabular}{cccccc}
\toprule
\textbf{Ratio} & \textbf{T} & \textbf{R1@0.5} & \textbf{R1@0.7} & \textbf{mIoU} \\
\midrule
1:1 & 32 & 68.52 & 47.29 & 58.16 \\
3:4 & 24 & 69.49 & 47.70 & 58.35 \\
1:2 & 16 & \textbf{70.65} & \textbf{49.58} & \textbf{59.78} \\
1:4 & 8  & 69.21 & 46.85 & 58.82 \\
\bottomrule
\end{tabular}
\vspace{-2mm}
\end{table}

\begin{table}[t]
\caption{Evaluation of Different Top-$k$ and $N$ Settings on QVHighlights Validation set.}
\label{tab:topk_n_settings_qvh}
\centering
\resizebox{\columnwidth}{!}{%
\begin{tabular}{cccccc}
\toprule
\textbf{N} & \textbf{Top-$k$} & \textbf{R1@0.5} & \textbf{R1@0.7} & \textbf{mAP@0.5} & \textbf{mAP@0.75} \\
\midrule
60 & 32 & 76.58 & 63.68 & 68.29 & 55.21 \\
60 & 50 & 76.19 & 63.10 & 69.20 & 55.27 \\
60 & 60 & 76.06 & 62.65 & 68.64 & 53.83 \\
80 & 32 & \textbf{76.97} & \textbf{63.74} & 68.40 & \textbf{55.30} \\
80 & 50 & 76.65 & 62.84 & 68.95 & 54.42 \\
80 & 60 & 76.52 & 62.45 & 68.53 & 54.02 \\
100 & 32 & 76.77 & 62.52 & \textbf{68.96} & 54.75 \\
100 & 40 & 76.58 & 63.03 & 68.63 & 55.67 \\
\bottomrule
\end{tabular}%
}
\vspace{-2mm}
\end{table}

\section{Post-Processing of LLM Outputs}
\label{sec:post-processing}

\begin{algorithm}[t]
\caption{Post-Processing Predicted Output}
\begin{algorithmic}[1]
\STATE \textbf{function} \texttt{post\_process(pred)}
\STATE pred $\gets$ clean input (e.g., remove extraneous markers)
\IF{pred is invalid or empty}
    \STATE \textbf{return} ``[[-1, -1]]"
\ENDIF
\STATE windows $\gets$ split pred into cleaned sub-windows
\STATE output $\gets$ empty list
\FOR{each window in windows}
    \STATE t\_start, t\_end $\gets$ extract and validate numbers
    \IF{t\_start \texttt{>} t\_end}
        \STATE swap t\_start and t\_end
    \ENDIF
    \STATE append \texttt{[t\_start, t\_end]} to output
\ENDFOR
\STATE \textbf{return} formatted output
\end{algorithmic}
\end{algorithm}

Despite training LLMs specifically for the moment retrieval task and instructing them to output results in the format of a nested list of moments:
\[
y = [[t_1^{\text{start}}, t_1^{\text{end}}], [t_2^{\text{start}}, t_2^{\text{end}}], \dots]
\]
LLMs can still produce outputs that deviate from the required structure due to inherent instability.
For instance, instead of the expected output \texttt{[[1.5, 4.3], [6.7, 9.2]]}, the LLM might generate malformed results such as \texttt{[[1.5, 4.3],[6.7, 9.2]} or \texttt{[[1.5, 4.3][6.7, 9.2]]}, causing errors during parsing. To address this issue, we implemented a robust post-processing mechanism to normalize the outputs and ensure strict adherence to the required format. This approach effectively mitigates formatting inconsistencies and supports reliable downstream processing.

The \texttt{post\_process} function refines predictions by removing extraneous markers ({\em e.g.}, ``\texttt{</s>}''), verifying compliance with the required nested list format, and correcting formatting issues.  If the output does not match the expected structure---where each element contains a start and end time---the function attempts to extract valid number pairs. It splits the output into individual moment windows using special symbols, cleans each window by removing extra commas, adjusts the format, and extracts start and end times. These values are validated, adjusted to valid indices, and swapped if necessary. The processed windows are then appended to the output list.

This post-processing pipeline ensures the final output adheres to the specified nested list format. Testing the model on over 50,000 samples confirmed the robustness of this approach, successfully normalizing all outputs for further analysis and moment retrieval tasks.

\FloatBarrier
\begin{figure*}[t] 
\centerline{
  \includegraphics[width=0.95\linewidth]{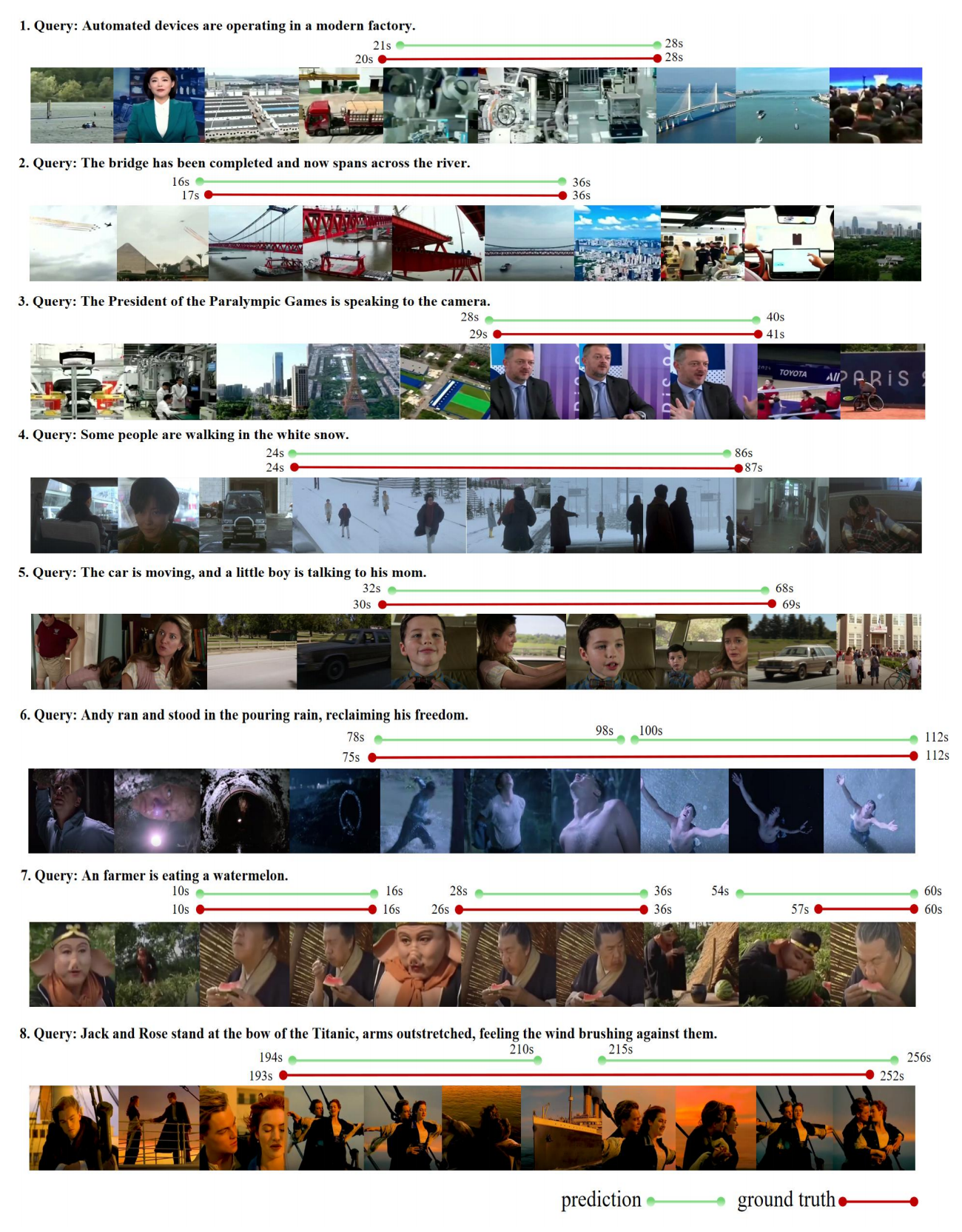} 
\vspace{-2mm}
}
\caption{Qualitative results across various types of videos, with ground truth segments for query events alongside highlighted predicted intervals.}
\label{fig:showcase_3} 
\vspace{-2mm}
\end{figure*}

\section{More Qualitative Results}
\label{sec:qualitative}

One of the key advantages of using MLLMs for moment retrieval tasks is their strong generalization and adaptability. Unlike transformer models with task-specific prediction heads, which are prone to annotation bias, MLLMs utilize cross-modal representations learned from large-scale multimodal data, enabling them to handle a wide range of video types effectively.

We demonstrated LLaVA-MR’s effectiveness on datasets like Charades-STA and QVHighlights, which feature relatively simple videos with minimal scene transitions. To further evaluate its versatility, we tested LLaVA-MR on more complex videos from diverse domains, including news clips and movies. These results, shown in Figure~\ref{fig:showcase_3}, highlight the model’s adaptability to intricate video styles and scenarios.

{\bf News Clips:}
We evaluated LLaVA-MR on challenging examples from news videos:

\begin{itemize}[leftmargin=10pt] \itemsep -.1em
\item \textbf{Example 1:} A fast-paced news flash with frequent scene transitions and multiple reports, where LLaVA-MR accurately retrieved a moment showing ``automated devices operating in a modern factory'', demonstrating its ability to accurately locate specific moments amidst diverse and rapidly changing visuals.

\item \textbf{Example 2:} A moderately paced news report on a bridge construction project. This coverage included various angles of the bridge and the construction process. The model accurately identified the relevant time intervals without missing any critical scenes. 

\item \textbf{Example 3:} A political report on the Paralympics, where LLaVA-MR correctly retrieved a segment featuring an interview with the Olympic Committee President. This result highlights LLaVA-MR's capability to handle complex queries, including those involving domain-specific terminology.

\end{itemize}

{\bf Movie Clips:}
We evaluated LLaVA-MR on movie clips, which present greater challenges for moment retrieval due to their intricate scene transitions and complex narrative structures compared to short YouTube videos.

\begin{itemize}[leftmargin=10pt] \itemsep -.1em
\item {\bf Example 4:} The model successfully captured the full context of a snowy scene, including the pivotal moment when the main characters meet and converse.

\item {\bf Example 5:} Despite the scene being filmed from multiple perspectives, depicting a boy talking with his mother in a car, LLaVA-MR accurately identified the complete interval.

\item {\bf Example 6:} In a scene where the male protagonist regains his freedom, the model divided the moment into two intervals, but their combined duration closely aligned with the ground truth.

\item {\bf Example 7:} The model mistakenly identified Zhu Bajie (a character from a Chinese classic) eating watermelon as a farmer eating watermelon, likely due to challenges in distinguishing the character's unique features.

\item {\bf Example 8:} While the model correctly identified key frames, it missed distant views, potentially due to resolution limitations affecting the detection of smaller visual elements.

\end{itemize}

These results highlight LLaVA-MR’s effectiveness but also suggest areas for improvement, such as increasing resolution and enhancing its ability to recognize fine-grained details.

\section{Future research potential}
\label{sec:Experiments}

The moment retrieval task has a wide range of applications, including video summarization, content-based retrieval, multimedia search, video recommendation systems, and educational video indexing. This section explores potential future directions for enhancing the performance of MLLMs in moment retrieval tasks.

A significant area for improvement is enabling MLLM-based models to output relevance scores for video clips with respect to a given query. Currently, LLaVA-MR is designed solely to predict the temporal intervals relevant to the query. However, unlike transformer-based methods equipped with prediction heads, LLaVA-MR lacks the capability to assign confidence scores to individual video clips. This limitation reduces its utility in applications that require score-based ranking or thresholding. Future research could focus on integrating MLLMs with prediction head structures, allowing the model to retain its strong contextual understanding for interval prediction while also generating relevance scores for each clip. Such an enhancement would bridge the gap between interpretability and robustness, significantly increasing the model’s versatility in real-world applications.

Expanding the inclusion of additional modalities in training and inference processes offers a promising direction for advancing moment retrieval tasks. Currently, our approach primarily relies on video frame sequences and textual queries, leaving other modalities underutilized.
On one hand, the information extracted from videos remains incomplete, as it focuses solely on visual content while neglecting the audio component. Developing effective methods to process and integrate audio alongside visual data is essential for enabling the model to reason holistically across both modalities. This integration could significantly enhance the model’s ability to capture richer and more nuanced contextual information.
On the other hand, the current prompt format is limited to text. Broadening this scope to include prompts in other formats, such as images or audio, could greatly expand the applicability of moment retrieval tasks. For example, an image prompt featuring a close-up of a bouquet of flowers could help identify the corresponding time segment where the bouquet appears in a video. Similarly, an audio prompt featuring cheering sounds could assist in locating scenes associated with such sounds. By accommodating diverse query formats, models could unlock new opportunities for moment retrieval in real-world applications, paving the way for more adaptable and robust solutions.
Furthermore, leveraging advanced strategies from LLMs, such as Chain-of-Thought (CoT) reasoning, Retrieval-Augmented Generation (RAG), and Reinforcement Learning (RL), presents promising pathways for enhancing MLLMs in moment retrieval tasks.Chain-of-Thought (CoT) reasoning involves breaking down complex reasoning processes into intermediate steps, enabling the model to address intricate tasks more effectively. In the context of moment retrieval, CoT could help the model analyze the relationships between time intervals in a sequential manner, thereby improving its ability to identify temporally complex events.
Additionally, Retrieval-Augmented Generation (RAG) can enhance model performance by incorporating external knowledge or context. For example, the model could use RAG to retrieve and integrate similar scenes or textual descriptions from external resources, improving localization accuracy. This approach could also help resolve ambiguities, such as distinguishing between visually similar objects in different contexts.
Moreover, Reinforcement Learning (RL) offers a framework for iterative, reward-based training, optimizing model performance over time. In moment retrieval tasks, rewards based on the alignment between predicted and actual intervals could refine the model’s temporal precision, ensuring better performance in tasks that require exact boundary identification.
Together, these strategies provide compelling opportunities to deepen, adapt, and strengthen the effectiveness of MLLMs, enabling them to tackle the unique challenges of moment retrieval with greater precision and versatility.